# The Analysis and Extraction of Structure from Organizational Charts

Nikhil Manali, David Doermann, and Mahesh Desai

Department of Computer Science and Engineering,
University at Buffalo

**Abstract.** Organizational charts also known as org charts are a crucial representation of an organization's structure and hierarchical relationships between its various components and positions. Extracting information from org charts manually can be time-consuming and error-prone. This paper presents an end-to-end and automated approach for extracting information from org charts using a combination of computer vision, deep learning, and natural language processing techniques. Furthermore, to evaluate the effectiveness of the information extracted, we propose a metric to measure completeness and hierarchical accuracy. This automated approach has the potential to improve organizational restructuring and resource utilization by providing a clear and concise representation of the organizational structure. This study serves as a foundation for further research on the topic of hierarchical chart analysis.

## 1 Introduction

An org chart is a visual representation of the internal structure of a company that illustrates the roles, responsibilities, and relationships of individuals within the organization. It is an effective tool for understanding the hierarchical structure of a business and the relationships between different departments and positions. Org charts are commonly used during restructuring or changes in management to help employees understand how their roles fit into the overall structure of the company.

One key aspect of org charts is their ability to be analyzed for further insight. To facilitate this, it is important to represent org charts in a format that is suitable for analysis. One such format is Visio, which includes parent-child relationships and detailed information about each node in tabular form. However, manually converting an org chart to a tabular format requires human involvement to specify the various nodes and their hierarchy, as well as to annotate the connections between parent and child nodes. Despite the many useful applications of org charts, there is currently a lack of tools that can automatically extract and store information from these charts in a suitable format for analysis. This highlights the need for further development in this area to improve the utility of org charts in business decision-making.

Therefore, we introduce an end-to-end approach to extract and analyze an org chart and represent it in a tabular format (see Figure 1). Our main contributions to this work can be summarized as follows: (i) Discuss and formulate a structural overview of the

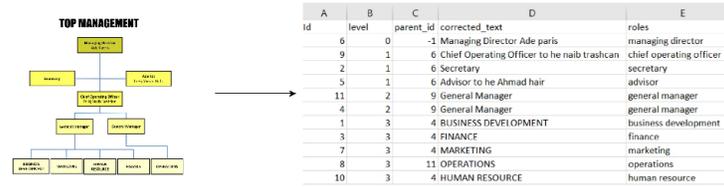

Fig. 1. Org chart and its tabular representation.

Org chart. (ii) End-to-end approach for automatic data extraction from an org chart. (iii) Structured the stored information in a format suitable for further analysis. (iv) Evaluation metric to measure completeness and hierarchical accuracy.

We also open-source our implementation and dataset for reproducibility and further research.

## 2    Related Work

### 2.1    Node Extraction and Topological Relationship.

Directed Graphs are visual representations of complex data that include text and various styles of nodes and edges, such as modeling graphs and flowcharts. Although these graphs are easily understood by humans, they pose a challenge for machines to interpret.

Some deep learning-based approaches are used to extract information from graphs and use this to obtain topological relationships. The Graph-Decoder[17] method employs a semantic segmentation network based on U-Net[12] as its backbone to extract data from raster images. The attention mechanism module is enhanced, the network model is simplified, and a specific loss function is designed to improve the model's ability to extract graph data. By using this improved semantic segmentation network, the Graph-Decoder method is capable of extracting the data of all nodes and edges, which are then combined to obtain the topological relationship of the entire Directed Network Graph. The proposed technique and Graph-Decoder method can be utilized to extract information such as text and topological relationships from Directed Network Graphs like Organizational charts.

A novel unsupervised method called Create (ChArT REcovEr)[19] for Organizational Chart Inference is proposed mainly for enterprise social networks (ESNs) charts. It consists of three steps for inferring organizational charts: Social stratification: ESN users are stratified into different social classes based on their social interactions. Supervision link inference: Supervision links are inferred from managers to subordinates. Consecutive social classes matching: Redundant supervision links are pruned by matching consecutive social classes. The authors evaluated Create on a real-world ESN dataset and showed that it can achieve state-of-the-art performance on the task of organizational chart inference. The paper also discusses the challenges of organizational chart inference and the limitations of Create. The authors conclude that Create is a promising approach for inferring organizational charts from ESNs, and that future work is needed to address the challenges of this task.

## 2.2 Graph Visualization:

Graph Visualization is important for analyzing and processing graphs. Some papers offer a systematic analysis of graph visualization, outlining common challenges and fundamental limitations. It includes a taxonomy scheme for general graph representations, an introductory discussion of critical constraints to be considered when developing a meaningful and useful visual graph representation, and a proposal for a visualization and exploration framework for graphs. **Visualizing Graphs - A Generalized View**[15] discuss i)taxonomy of network presentation techniques ii)the different constraints to be considered, when visualizing complex structures[1]. Finally, how the different approaches can be combined in a framework for exploring complex graphs.

This paper describes and discusses the taxonomy of visualization methods for hierarchies and also introduced its taxonomy for network visualization techniques. Two principal alternatives[15] to classify visualization techniques for hierarchies:
- explicit vs. implicit
- axes-oriented vs. radial

1. **Explicit vs. Implicit:** Explicit methods utilize a traditional node-link representation to display the connections between hierarchy elements. In contrast, implicit techniques use space-filling methods and a diverse range of abstract node representations, such as lines, boxes, and circles, to depict relationships between elements.
2. **Network Representations:** we propose three possible categorizations for visualization techniques for networks:
   - directed vs. undirected
   - explicit vs. implicit
   - free, styled, or fixed

   Furthermore, they describe various visualization-related constraints like User Constraints, Data Constraints, Aesthetic Constraints, etc. Graph visualizations can be used to communicate information efficiently, hence a thorough examination of graph visualization, particularly in the case of Flow charts, such as Org charts, aids in the proposition of effective data extraction methods.

## 2.3 Chart mining:

Chart mining refers to the automated process of detecting, extracting, and analyzing charts to retrieve the tabular data used to create them. This process enables access to data that may not be available in other formats and facilitates the development of various downstream applications. [3]The different methods used throughout the automated chart mining process includes: i)the automated extraction of charts from documents, ii) the handling of multi-panel charts, iii)the use of automatic image classifiers to collect chart images on a large scale, and iv) the automated extraction of data from different chart types.

# 3 Dataset

We collected more than 2500 org chart images from various sources, mainly online. To supplement the limited number of available samples, we augmented the original images and also generated synthetic images that mimic the structure of the org chart with varying shapes and text.

**Synthetic Org charts:** The organizational chart comprises nodes of distinct shapes, each containing text describing them. We applied computer vision techniques to generate synthetic org charts by generating multiple shapes(like rectangles, squares, and eclipses) and added text within them. Subsequently, we arranged these shapes in a structured layout with rows and columns, replicating the standard format of an organizational chart. These synthetic org charts are used to increase the training dataset for node detection tasks.

We used shape transformation and HSV color transformation techniques for augmentation to generate four augmented samples from a single image. As no annotated data were available and annotating an org chart is a tedious task, we developed a semi-automatic data annotation pipeline. **First**, we used computer vision techniques to detect nodes using structural kernels along with their bounding boxes. The inaccurate predictions were then filtered out and we were able to collect 1592 images which were then augmented with different augmentation techniques to increase the dataset. We also generated 500 synthetic images with bounding boxes to make our dataset more diverse and to avoid overfitting. A total of 4590 samples were used to train our object detection models.

The dataset is formatted similarly to the COCO dataset[8], given that we have leveraged a pre-trained model on the COCO dataset. After training, only the original organizational chart images were used for subsequent steps in the pipeline

## 4  Our Approach

The analysis of an org chart requires the extraction and tabulation to be both efficient and meaningful. Our methodology involves a multistep process that efficiently extracts information from an org chart. To perform a meaningful analysis of an organizational chart, it is crucial to have a thorough understanding of its structure and key components. In section 4.1, we have highlighted the vital components within the organizational chart and represent its structure in a way that streamlines the process of working with the chart.

### 4.1  Structural overview

The diagrammatic representation of an org chart serves to visually communicate the internal structure of a company, depicting the roles, responsibilities, and relationships between individuals within the entity. These diagrams use simple graphical symbols, such as lines, squares, and circles, to connect related job titles and positions within an organization. The structure of an org chart can be divided into three distinct components:

1. **Nodes or Blocks:** Each node/block represents and visually communicates a company's internal structure, the person's name, role within the organization, and possibly an image. (Figure 2)
2. **Connection Lines/Edges:** The relationships between nodes in an org chart are represented by lines connecting one node to another node(parent nodes to child nodes), which can be direct or intersect to form a junction. These connecting lines can be interpreted as directed edges from the parent node to the child node. In

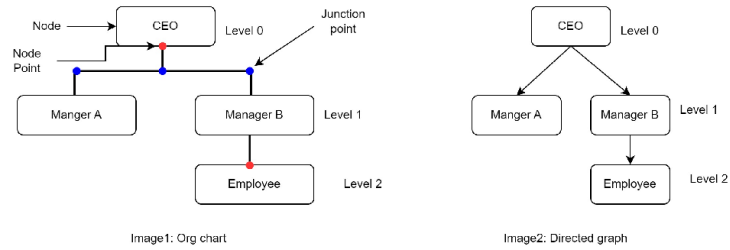

**Fig. 2.** org chart structural description

Figure 2 ( See Image 2), a connecting line between *2* levels of the node where $level_0 > level_1$ in a hierarchy is represented as a directed edge.

3. **Node and junction points:** Node points are formed at the intersections of a node and an edge, while junction points are the intersections of two edges.
   The utilization of these structural components is a crucial aspect of our approach in the subsequent stages of our work.

The process of extracting information from an org chart or any other hierarchical chart involves the following two key steps:

1. Identification of individual nodes or data blocks.
2. Analysis of the relationships between these nodes and other related nodes.

### 4.2 Localization of Node or Data blocks

Localizing or detecting nodes is crucial to determine different levels and positions in an org chart. A node, as described in Section 4.1, holds information about an individual and its position within the organization. To extract information accurately, the location of the node and its contents must be identified.

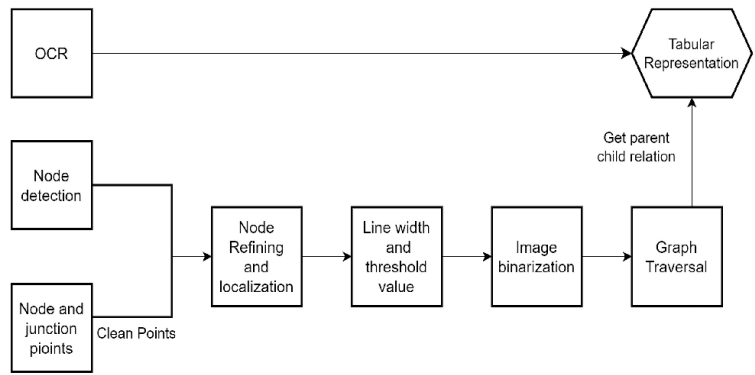

**Fig. 3.** org chart Analysis Pipeline

**Node location:** We approached this task as an object detection problem, where the nodes in an organizational chart were considered as objects with shapes such as squares, rectangles, and ellipses. We used three different state-of-the-art object detection models: Yolo[10], CornerNet[7], and DETR[2]. We used Yolo as a baseline model, and then chose DETR for this task because it had a higher mean average precision (mAP) than the other two models as shown in the table 1 and its superior performance compared to other anchor-based models like YOLO. Our training pipeline is semi-supervised and involves the automatic labeling of data (see Section. 3) using computer vision techniques to detect shapes. These labeled images were then used as training examples for our object detection models.

We further enhanced the localization of the bounding boxes from the output by implementing corner refinement. The bounding box is defined by its corner point coordinates $(x, y)$ and dimensions (height (H) and width (W)) as $(x, y, h, w)$ or $(x_1, y_1, x_2, y_2)$, where $(x_1, y_1)$ and $(x_2, y_2)$ represent the upper left and bottom right corners, respectively. We created a region of interest around the DETR bounding box corner and utilized the Harris corner detector to obtain precise corner points. The corner points obtained were then used to calculate new values $(x_1, y_1)$ and $(x_2, y_2)$, resulting in a refined bounding box.

**Text detection:** Once the precise locations of the nodes were obtained, we proceeded to extract the information contained within them. The information was in the form of text and, therefore, we used optical character recognition (OCR) to detect it. We use easy-OCR[5] on the images to retrieve the text and its location and then map this information to the respective node.

### 4.3 Analyzing and Finding relation between nodes

The analysis and identification of the relationship between nodes is a critical aspect of understanding the hierarchical structure of an org chart. Once the precise location of the nodes and the corresponding text have been established, the next step involves determining the connections and relations between the various nodes. The accuracy of these relationships is paramount in presenting the org chart in a structured tabular format.

The org chart can be represented as a directed graph, with directed edges connecting the parent node to its child node (see Figure 2 in Image 2). To establish these relationships, we used the depth-first search (DFS) for the traversal of the graph and the topological ordering of the nodes. This allowed us to determine the parent-child relationships and their respective levels in the graph hierarchy, with the root node being at level one, its children at level two, etc.

To properly order the directed graph, it was necessary to identify the root node that holds the highest position in the organization. However, there are several challenges associated with finding the root node in an org chart. Specifically, the top node may not always be the root node if the org chart is constructed from the left or right side. To address this challenge, we compiled a corpus of all the available positions in the organization and their corresponding seniority levels. The corpus included more than 500 positions categorized by organization and seniority level, where a score of 1.0 indicates the most senior position.

To determine the root node of a given org chart, we followed the following steps:

1. Identify all positions in the org chart.

2. Traverse the corpus and obtain the seniority score of each position.
3. The root node is the position with the highest score which corresponds to the highest seniority level.

In many cases, the most senior position in an organization chart may not be the highest position. For example, in a corporation, the Board of Directors is the highest position, but some org charts start with the CEO. In this case, the CEO position will receive the highest score among all positions available in the org chart and be considered the root node. This approach enabled us to accurately predict the highest seniority position or the root node.

**Node and Junction points:** As mentioned in section. 4.1, the Harris corner detector[6] is used to find the intersection points in a graph, which are then classified into node and junction points. We also used Shi-Tomasi[16] *goodFeaturesToTrack()* to refine the corner points.

Instead of $R = Scoring\ Function$ in Harris Corner

$$R = \lambda_1 \lambda_2 - K(\lambda_1 + \lambda_2)^2$$

we use:

$$R = min(\lambda_1, \lambda_2)$$

1. **Node Points:** A node point is a corner point where a node intersects a connection line or edge and is classified as a Node point (denoted with a red dot in Figure 2).
2. **Junction points:** Junction points are points at the intersection of two edges.

   Junction points are important because they reveal the characteristics of the connection lines or edges at their intersections. To determine which pixels in an image belong to these lines/edges, it is necessary to correctly classify corner points as either node points or junction points. Points that are not classified as node points are considered junction points. However, many unwanted junction points may be detected due to the presence of different shapes in the background of the org chart. To eliminate these unwanted junction points, a bounding box is formed around potential junction points and if the points do not result from the described intersection as shown in Figure 4, they are filtered as invalid junction points.

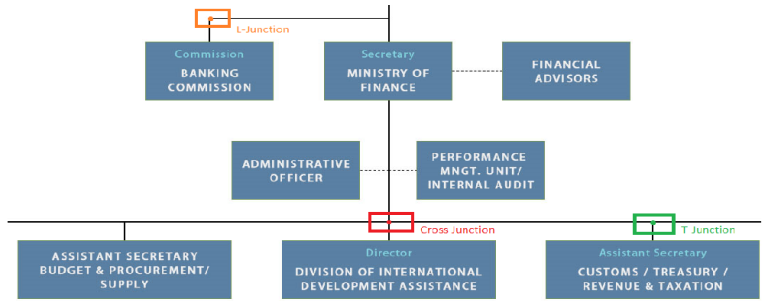

**Fig. 4.** Points at the cross, L and T junctions.

8    Nikhil Manali, David Doermann, and Mahesh Desai**Image Thresholding:** Although we can visually identify edges between our nodes, it is important to have a clear and precise classification of edges at the pixel level for graph traversal. To obtain this information, we utilize junction points, as they are formed at the intersection of connection lines or edges, and provide valuable information at the pixel level. We used all these junction points as a center and created a bounding box around them. Then, OTSU[9][18] thresholding (Figure. 5) was performed in these sections, generating a histogram representing the distribution of pixel intensities.

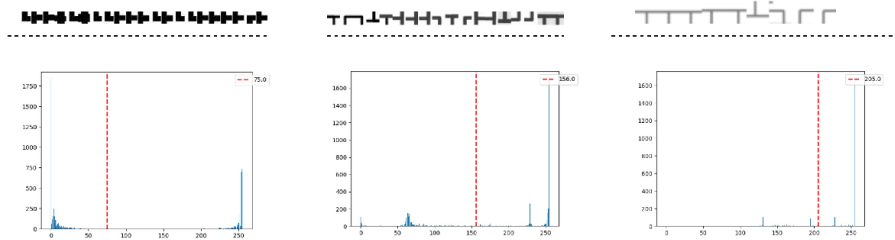

**Fig. 5.** A few examples of junction point regions that are used to find the threshold value using OTSU thresholding. The red lines in the figure show the division of pixel densities into two clusters, and this line represents the threshold value.

The core idea behind OTSU thresholding is to separate the image histogram into two groups with a threshold defined as a result of minimizing the weighted variance of these classes denoted by $\sigma_w^2(t)$.

The whole computation equation can be described as follows.

$$\sigma_w^2(t) = w_1(t)\sigma_1^2(t) + w_2(t)\sigma_2^2(t)$$

where $w_1(t), w_2(t)$ are the probabilities of the two classes divided by a threshold t, ranging from 0 to 255 inclusively.

The probability P is calculated for each value of the pixels in two separate groups $C_1, C_2$ using the cluster probability functions expressed as:

$$w_1(t) = \sum_{i=1}^{t} P(i), w_2(t) = \sum_{i=t+1}^{I} P(i)$$

After applying OTSU thresholding, the image was binarized[9] into foreground and background segments. The foreground was represented by pixels with a value of 0, while the background was represented by pixels with a value of 255.

**Graph Traversal:** At this point, we have also established a connected graph consisting of nodes and edges, and using the thresholding technique, as outlined in Section 4.3, we have identified the pixels that belong to the edges. To facilitate the traversal of the graph, we initiate the process from the root node and assign parent-child relationships to each node, while also determining their hierarchical level within the graph.

The root node is assigned a *level* → 0. To facilitate traversal, node points are treated as vertices rather than nodes, since they can be localized using coordinates (x,y) in an image.

As mentioned in section 4.3 we traverse the graph using Depth First Search (Pseudo Code 1) and also perform topological ordering of the nodes that serve as a hierarchy in a graph with root node as a top level. This process involves finding all nodes connected to the root node, assigning child IDs, and determining their hierarchy level within the graph. The time complexity for the whole operation is $O(V + E)$ (V is a vertex or node point and E is Edge).

---

**Algorithm 1** Graph Traversal
---
**Require:** Binary Image
**Ensure:** $Edges = 0px$
  $currParent \leftarrow root$
  $Edges \leftarrow 0$ px
  $NP \leftarrow nodePoint$  // color code Id
  **while** (x,y) is Edge **do**
    **if** $(x, y)$ is NP **then**
      $Parent(x, y) \leftarrow currParent$
      $Graph(currparent).add((x, y)$
      $currParent \leftarrow (x, y)$
    **end if**
    $visit(x, y) \leftarrow True$
  **end while**
  $Graph \leftarrow return$

---

## 5 Evaluation metrics

Our model generates a tabular representation, including the nodes, text within nodes, and their internal relationships with each other. Our tabular data format is similar to the standard Visio format with a few simple modifications. Visio is used to create diagrams like flowcharts and org charts and represents all of the essential information we extract. To assess the performance of the model, we require a suitable evaluation metric. We view org charts as directed graphs with edges connecting the root node to its children in a hierarchical manner. One commonly used metric for evaluating the similarity of graphs is the Graph Edit Distance (GED)[14] [4]. GED measures similarity (or dissimilarity) between two graphs. This metric has numerous applications, including error-tolerant pattern recognition in machine learning. The graph edit distance can be related to the string edit distance, with the interpretation of strings as connected, directed, acyclic graphs of maximum degree one. The graph edit distance between two graphs $g1$ and $g2$, written as $GED(g1, g2)$ can be defined as:

GED can capture the structural similarity of two graphs, but we also need the accuracy of the information within the node. We decided to evaluate the graph using a different approach, centered on the amount of information we can extract from the graph. This information can be divided into two parts: (1) The localization of nodes

$$GED(g_1, g_2) = \min_{(e_1,...,e_k) \in \mathcal{P}(g_1,g_2)} \sum_{i=1}^{k} c(e_i)$$

where $\mathcal{P}(g_1, g_2)$ denotes the set of edit paths transforming $g_1$ into $g_2$ and $c(e) \geq 0$ is the cost of each graph edit operation $e$ and accurate extraction of information from those nodes, and (2) the structural accuracy of the graph, including correct identification of parent-child relationships.

The ground-truth and predicted (Y) graphs are in tabular format and have IDs for nodes, so our first step is to match nodes from the ground-truth to the Y graph. To achieve this, we use cosine similarity[13] scores to match text information within nodes instead of Intersection over Union (IOU)[11] because it provides a clearer indication of which nodes match and how accurately we have extracted information from those nodes.

1. **Node Similarity** ($N_S$): This score provides the accuracy of node prediction by measuring the degree of similarity between the predicted nodes (Y) and the ground truth nodes (GT). The evaluation process involves selecting a node and determining its presence in the predicted nodes using cosine similarity. The GT node is assigned to its corresponding Y node if the cosine similarity[13] score $S_C$ exceeds 0.95. Here, Cosine Similarity

$$S_C(A, B) := \cos(\theta) = \frac{\mathbf{A} \cdot \mathbf{B}}{\|\mathbf{A}\|\|\mathbf{B}\|} = \frac{\sum_{i=1}^{n} A_i B_i}{\sqrt{\sum_{i=1}^{n} A_i^2} \sqrt{\sum_{i=1}^{n} B_i^2}},$$

Here A and B are text in the GT node and Text in Y Node. The highest score we can get is $n$ *(number of nodes)*, if we predict all nodes correctly.

2. **Structural accuracy** ($S_A$): Having successfully mapped the node IDs between the ground truth (GT) and the predicted graph (Y), the next step is to determine the structural accuracy of the predicted graph. This involves evaluating the accuracy of predicting parent-child relationships, which are represented as directed edges from node "$u$" to node "$v$" in the graph.

$$Graph(parent \to p, child \to c) : p \to c$$

The total score $T_S$ is calculated as the average of the node similarity and the structural accuracy:

$$\frac{\sum_{i=1}^{n}(N_S + S_A)}{n} ; n = \text{no of nodes.}$$

The total score ranges from $T_s \in (0.0, 1.0)$

## 6  Experiment

We perform experiments with different stages, including multiple deep learning models for node detection, various thresholding techniques for image binarization, and filtering false corner points.

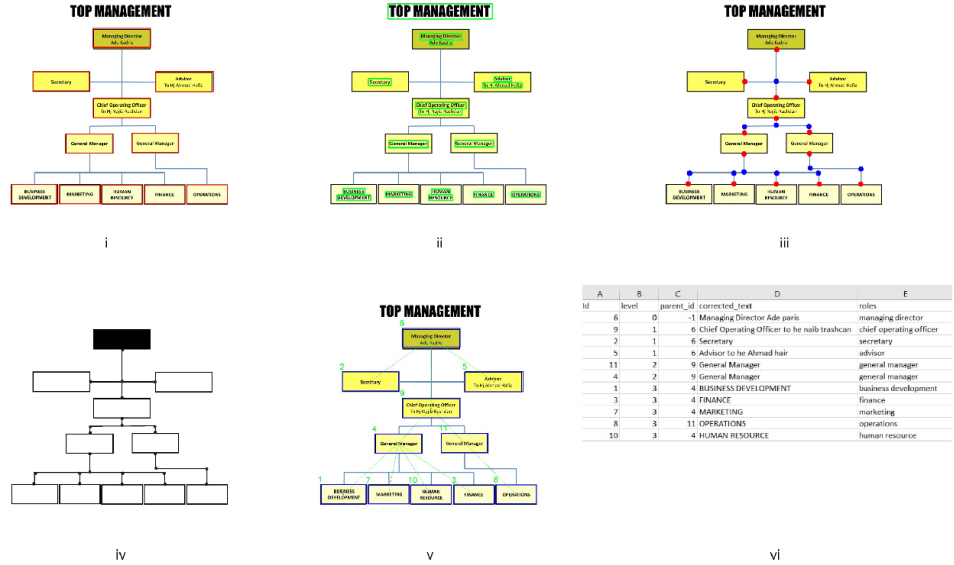

**Fig. 6.** (i) Node Detection. (ii) Text Detection. (iii) Node and Junction points. (iv) Image Binarization. (v) Graph. (vi) Data extracted in tabular format.

### 6.1  Node and Text detection

For our object detection task, we employed three distinct Deep Learning models: YOLO, CornerNet, and DETR. These models were pre-trained on the MS-COCO dataset and further fine-tuned on the same dataset samples, with similar learning rates and three different epoch configurations (25, 50, and 100 epochs). The performance of the models is evaluated based on Mean Average Precision (mAP) and Mean Average Recall (mAR) score.

Initially, YOLO was selected as a baseline due to its speed and comparable accuracy with similar configuration models. However, YOLO's performance was unsatisfactory, particularly for smaller boxes in the organizational chart, and required extensive post-processing for fine-tuning and proper localization of bounding boxes around nodes. To address localization problems, we opted for the CornerNet model, which detects an object bounding box as a pair of key points (top-left and bottom-right corners) using a single convolution neural network. By detecting objects as paired key points,

CornerNet eliminates the need for designing a set of anchor boxes commonly used in prior single-stage detectors(Like Yolo), and the corner pooling layer helps the network better localize corners. CornerNet was able to localize well around bounding boxes, but it detected many false positives.

To address these challenges,(i). the BBox localization and,(ii) reducing false positives we used the Facebook-developed detection transformer model (DETR), which excels at end-to-end detection. This DETR model was pre-trained on the COCO dataset and further fine-tuned on our proprietary dataset, with an evaluation conducted on 236 samples. DETR is a state-of-the-art object detection algorithm that uses a transformer-based architecture and one of the key benefits of using DETR is that it eliminates the need for anchor boxes and non-maximum suppression (NMS), and hence improved localization of Bounding Boxes as compared to YOLO.

Table 1. Model Comparison by AP(Average Precision), mAP(mean Average Precision) and mAR Accuracy(mean Average Recall)

| Models | AP | mAP Score | | mAR Score | |
| --- | --- | --- | --- | --- | --- |
| | | $mAP_{50}$ | $mAP_{75}$ | $mAR_{50}$ | $mAR_{75}$ |
| YOLO V3 | 0.8439735 | 0.18026860 | 0.1302712 | 0.9131932 | 0.84398143 |
| CornerNet | 0.91323 | 0.22345 | 0.17865 | 0.921103 | 0.89711 |
| DETR | 0.92658 | 0.2483 | 0.2154 | 0.92659 | 0.920897 |

For text detection, we utilized a state-of-the-art Optical Character Recognition (OCR) tool called easy-OCR. All the text detected inside a node was then merged into a single entity.

### 6.2   Node and Junction points

Node and junction points are a subset of corner points and Initially, we used the Harris corner[6] detector for feature detection; however, the resulting points lacked refinement. To improve the localization of these points, we used the Shi-Tomasi[16] Good Features to Track method.

### 6.3   Thresholding

We explored various thresholding methods to accurately extract org charts from images. Initially, we tested normal thresholding on the charts but found that this approach was not effective since we could not use a single global threshold value for multiple org chart images.

Subsequently, we investigated Adaptive thresholding, which consists of two types: Mean Adaptive Thresholding and Gaussian Adaptive Thresholding. Adaptive thresholding computes the threshold for each pixel based on a small region surrounding it, generating different thresholds for different parts of the same image. This method was more effective than normal thresholding, as the threshold varied for each small part of the image, rather than being constant for entire images and all images. However, we still could not achieve the desired org chart structure using Adaptive Thresholding as it applied thresholding within a small bounding region in the image, which could overlap with the org chart (foreground) and the background. The thresholding generated for each pixel affected the foreground as well, making it difficult to separate the foreground from the background. Correct thresholding is very important for graph traversal as mentioned in 4.3 and to address this issue, we utilized Otsu's Threshold method, which calculates the threshold value based on the image's histogram. This threshold value distinguishes the foreground from the background and significantly improved our results see Fig. 7.

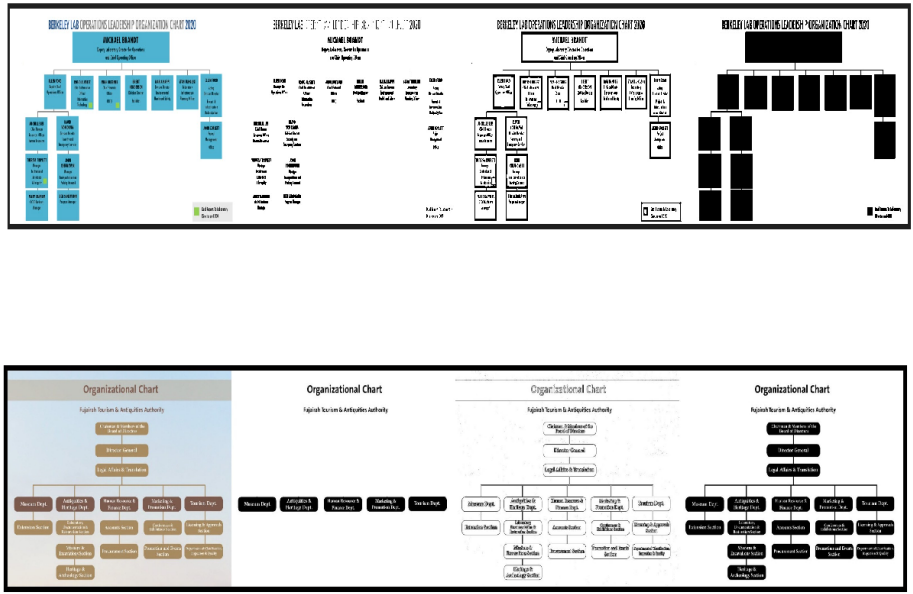

**Fig. 7.** (i) Original Image. (ii) Normal Thresholding. (iii) Adaptive Thresholding. (iv) OTSU Thresholding.

Next, we aimed to obtain only the org chart by majorly considering the foreground and less of the background. We decided to select specific points present on the org chart itself, such as junction points, to improve the threshold value. Finally, we applied Otsu Thresholding on the combined image of junction points, which resulted in a good threshold value that was closer to the foreground and eliminated most of the background. Consequently, we obtained a proper threshold image, which contained only the org chart.

## 7   Results

For the analysis of our end-to-end org chart model, we have evaluated the final tabular output using 120 annotated samples. The ground truth file includes essential information, such as the node ID, parent ID, and the information available within each node.

As described in Section 5, we have employed our custom metric to evaluate the accuracy of the information contained within the graph, including the location and content of the node, and the structural precision of the graph itself.

The presented results are in the Table. 7 are based on the results obtained from three distinct models, namely YOLO, CornerNet, and DETR. The OTSU thresholding method was employed for binary thresholding and DFS(Depth First Search) for graph traversal. The DETR model with Otsu thresholding outperforms the other models, as demonstrated in Table 7. On the other hand, CornerNet exhibited a relatively poor performance compared to YOLO and DETR due to its high rate of false positive BBox detections. These false positive detections are difficult to eliminate after post-processing refining steps, which consequently reduces the accuracy of OTSU thresholding. It should be noted that accurate Bounding Boxes(see section 6.3) play a vital role in determining the correct thresholding value.

Table 2. Node and Structural Accuracy Results.

| Model | Mean $N_S$ | Mean Structural | Median $N_S$ | Median Structural | Total score |
|---|---|---|---|---|---|
| YOLO | 0.71653 | 0.5120 | 0.73267 | 0.5287 | 0.614265 |
| CornerNet | 0.6824 | 0.432117 | 0.74897 | 0.4798 | 0.5572585 |
| Detr | 0.86274 | 0.530956 | 0.93218 | 0.588235 | 0.696851 |

As can be seen in Table 7, we obtained relatively low structural scores, which can be attributed to the difficulty in obtaining the correct thresholding value for certain org charts due to noise in the images. An incorrect threshold value can result in missing connections, which in turn hinders complete graph traversal, as there is no way to reach a child node with a broken connection

## 8   Analysis and Future.

Our approach to analyzing and extracting data from organizational charts has demonstrated remarkable accuracy and improvement compared to the baseline model as shown in the Table 7. We have achieved considerable improvement as compare to our baseline approach by utilizing advanced techniques to enhance text detection, corner detection, and thresholding values for binarization. However, there are several areas that necessitate improvement in order to optimize the overall system performance. These areas include:

1. Presently, our primary focus is on improving the efficiency of the text extraction process from low-quality images. Although our current method 4.2, which employs easy-OCR, has proven effective, it does have certain limitations, such as misspelled words in the case of low-resolution images. To overcome this issue, we intend to incorporate computer vision techniques to enhance the accuracy of text extraction from such images.
2. We aim to refine the node and junction points to ensure the accurate representation of the graph. Identifying these points accurately is critical for determining the appropriate threshold value. By enhancing our ability to identify these points, we anticipate a boost in our Structural Accuracy.
3. We have observed that our structural accuracy is affected by the existence of multiple parents for the same node. Our current approach is designed to handle only one parent for each child node. However, we plan to direct our future improvement efforts towards addressing this limitation and enabling the proper handling of such situations.

# 9 Conclusion

We presented a novel approach for the analysis and data extraction of organizational structure using deep learning and computer vision techniques. Our approach offers significant accuracy in extracting both text and structural information, which is then stored in a suitable format for further analysis. We have also developed a metric to evaluate the results of our approach. This methodology can be applied to other hierarchical charts such as flowcharts, however, it also presents certain challenges, particularly in detecting nodes and junction points, determining appropriate thresholds, and enhancing structural accuracy.